\documentclass[conference]{IEEEtran}
\IEEEoverridecommandlockouts
\usepackage{cite}
\usepackage{amsmath,amssymb,amsfonts}
\usepackage{algorithmic}
\usepackage{graphicx}
\usepackage{textcomp}
\usepackage{xcolor}
\usepackage[bookmarks=true]{hyperref}
\usepackage{bookmark}
\def\BibTeX{{\rm B\kern-.05em{\sc i\kern-.025em b}\kern-.08em
    T\kern-.1667em\lower.7ex\hbox{E}\kern-.125emX}}
\begin{document}

\title{Design of magnetic coupling-based anti-biofouling mechanism for underwater optical sensors\\
\thanks{This work was funded by the EU H2020 FETOPEN Project ‘‘Robocoenosis—ROBOts in cooperation with a bioCOENOSIS’’ [grant agreement no. 899520], and by the project “Heterogeneous Swarm of Underwater Autonomous Vehicles” [contract no. TII/ARRC/2047/2020]. The funders had no role in the study design, data collection and analysis, decision to publish, or preparation of the manuscript. }
}
\author{\IEEEauthorblockN{Jane Pauline Ramirez\IEEEauthorrefmark{1}, Cesare Stefanini\IEEEauthorrefmark{2}, Giulia De Masi\IEEEauthorrefmark{3}, and Donato Romano\IEEEauthorrefmark{4} }\IEEEauthorblockA{\IEEEauthorrefmark{1}\IEEEauthorrefmark{2}\IEEEauthorrefmark{4}The BioRobotics Institute, Sant'Anna School of Advanced Studies, Viale Rinaldo Piaggio 34, 56025 Pontedera, Pisa,  Italy}\IEEEauthorblockA{\IEEEauthorrefmark{2}\IEEEauthorrefmark{4}Department of Excellence in Robotics and AI, Sant'Anna School of Advanced Studies, 56127 Pisa, Italy}\IEEEauthorblockA{\IEEEauthorrefmark{2}Healthcare Engineering Innovation Center, Khalifa University, Abu Dhabi, UAE}\IEEEauthorblockA{\IEEEauthorrefmark{3}Autonomous and Robotics Research Centre, Technology Innovation Institute, Abu Dhabi, UAE}\IEEEauthorblockA{\IEEEauthorrefmark{3}College of Engineering, Khalifa University, Abu Dhabi, UAE}\IEEEauthorblockA{Email: \IEEEauthorrefmark{1}ramirezjanepauline@gmail.com, \IEEEauthorrefmark{2}cesare.stefanini@santannapisa.it, \IEEEauthorrefmark{3}giulia.demasi@tii.ae, \IEEEauthorrefmark{4}donato.romano@santannapisa.it}}

\maketitle
\pdfbookmark[section]{Abstract}{abstract}

\begin{abstract}
Water monitoring is crucial for environmental monitoring, transportation, energy and telecommunication. One of the main problems in aquatic environmental monitoring is biofouling. The simplest method among the current antifouling strategies is the use of wiper technologies like brushes and wipers which apply mechanical pressure. In designing built-in strategies however, manufacturers usually build the sensor around the biofouling system. The current state-of-the-art is a fully integrated central wiper in the sensor that enables cleaning of all probes mounted on the sonde. Improvements in antifouling strategies lag rapid advancements in sensor technologies such as in miniaturization, specialization, and costs. Hence, improving built-in designs by decreasing size and complexity will decrease maintenance and overall costs.

This design is targeted for the EU project Robocoenosis since bio-hybrid systems in this project incorporate living organisms. This technology targets selective proliferation of the organisms which only prevents biofilms on components where they are unwanted. Beyond this, the use of autonomous activation based on image processing may likely be advantageous for minimizing the need for human inspection and maintenance. In addition to Robocoenosis, we also aim at incorporating this design in another project entitled Heterogeneous Swarm of Underwater Autonomous Vehicles where a swarm of heterogeneous underwater robotic fish is being developed.

\end{abstract}


\section{Introduction}
Biofouling is the undesirable surface biological adhesion and growth \cite{b7}. This is one of the main problems in aquatic environment monitoring \cite{b6}. The Alliance for Coastal Technologies estimated that half of the budget for operation is due to biofouling depending on location and season \cite{b1}. Disadvantages include shortened usable life of structures and instruments, increased energy consumption and worsens functioning and data quality usually causing sensor reading drifts \cite{b3}. Industries like transportation, renewable energy, telecommunications, and ocean and coastal observation are negatively affected \cite{b2}.

Biofouling occurs in four stages: (1) Adsorption of both inorganic and organic molecules as a film, (2) Reversible settlement of bacteria called microfouling (3) Irreversible development of more complex community called macrofouling, and (4) surface fixation of larger marine invertebrate \cite{b10}. 

Current antifouling strategies for sensor systems include wiper technologies, biocide generation systems, antifouling coatings, use of electrochemistry and irradiation. Wiper technology is deemed as the simplest method to remove biofouling. It is a mechanical pressure cleaning method using brushes and wipers \cite{b5}. Main considerations for using these types of systems are the use of materials, abrasion and maintenance which involves replacement and removal of components. Additionally, power requirements, housing ports, sealing, corrosion, and water intrusion. 

Stand-alone options are expensive, somewhat bulky as it houses actuators which requires more consideration in the design in terms of power, size, and housing. In designing built-in strategies, manufacturers usually build the sensor around the antifouling system \cite{b4}. The current state-of-the-art is a fully integrated central wiper in the sensor that enables cleaning of all probes mounted on the sonde. Improvements in antifouling strategies lag rapid advancements in sensor technologies such as in miniaturization, specialization, and costs\cite{b4}. Hence, improving built-in designs by decreasing size and complexity will decrease maintenance and overall costs.  

This paper aimed to construct, fabricate, and test the effectiveness of a magnetic coupling-based mechanism for antibiofouling. In Section \ref{matmet} Design and Methodology, we discussed the design of the built-in electromechanical wiper, the camera module and sensor unit, the experimental design, and the procedure for data analysis. This was followed by the sections on findings, conclusions, and recommendations for future study.

We infer that the use of an antifouling wiper mechanism based on magnetic coupling can be an effective component when incorporated in the existing designs today. However, more experiments are needed to validate the following prospective: less design issues are to be considered as compared to some build-in wipers because most of the moving parts will be housed inside the sensor compartment and are not exposed outside, less power is needed to drive the motor as compared to linear actuators used in stand-alone mechanical technologies because less components are needed to be driven, and less material and design issues will arise because the magnetic brush or foam will be the only one exposed to the environment which can be easily replaced as necessary.

The target application is a project of European Fostering Novel Ideas ‘FET Open’ Horizon 2020 project: ROBOts in cooperation with a bioCOENOSIS (Robocoenosis) which seeks to introduce a novel paradigm of 'life form in the loop'. This develops autonomous, robots and non-invasive bio-hybrid systems by using symbiotic methods to integrate both well-adapted living organisms in the environment, biodegradable materials, and technologies for energy harvesting, sensing and actuation \cite{b8}. A mechanism driven by gears and lead screws was placed inside the sensor compartment which will be coupled to a magnet on the surface of image sensors to get rid of building up biofilms. Since Project Robocoenosis bio-hybrid systems incorporate living systems, this technology targets selective proliferation of the organisms which only prevents biofilms on components where they are unwanted. This hypothesis was validated by an experiment assessing the image quality of the sensor samples. The use of the mechanism disrupted the biofilm growth in the wiping area coverage.

In addition to Project Robocoenosis, we are looking into the possibility to incorporate this design in another project entitled Heterogeneous Swarm of Underwater Autonomous Vehicles (Contract No. TII/ARRC/2047/2020). In this research project, a swarm of heterogeneous underwater robotic fish is developed. Starting from the biological counterparts of a real school of fishes, an artificial school of 30 hybrid (remotely controlled and autonomous) underwater robots is created. This consists of five special fish integrated with more sensors and communication channels with the remote operator, and 25 normal autonomous fish, with fewer sensors and are able to communicate each other. Additionally, a floating beacon can collect data from the fishes. A static robot deployed on the sea floor with the ability to resurface after the end of the operation is designed to collect environmental data. The main objectives are high fidelity teleoperation in underwater wireless robotics, especially focused on the five special robotic fishes, and collaborative tasks, collective motion and collective decision making in a school of artificial fishes using short range (optical modem and electric sense) and long-range communication channels (acoustic modem). These studies also include numerical simulations (Multi-agents and ROS/Gazebo), mathematical models, prototyping and experimental testing.

\section{Design and Methodology} \label{matmet}
This section describes the process towards designing the final prototype. Diagrams and figures such as the electrical diagram, mechanical models, and flowcharts, were included to augment the explanations. Also, manufacturers, part numbers and dimensions of different components were indicated to ensure the applicability of the design. 

\subsection{Mechanical Design}

The idea was to make a sliding magnetic mechanism coupled before and after the observation screen, which separates the sensor and the underwater environment. We have tested and fabricated two mechanisms namely the scotch yoke and lead screw.

\begin{figure}[t]
\centerline{\includegraphics[width=3in]{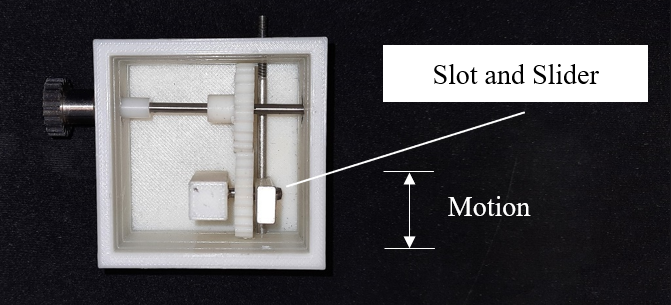}}
\centerline{\includegraphics[width=3in]{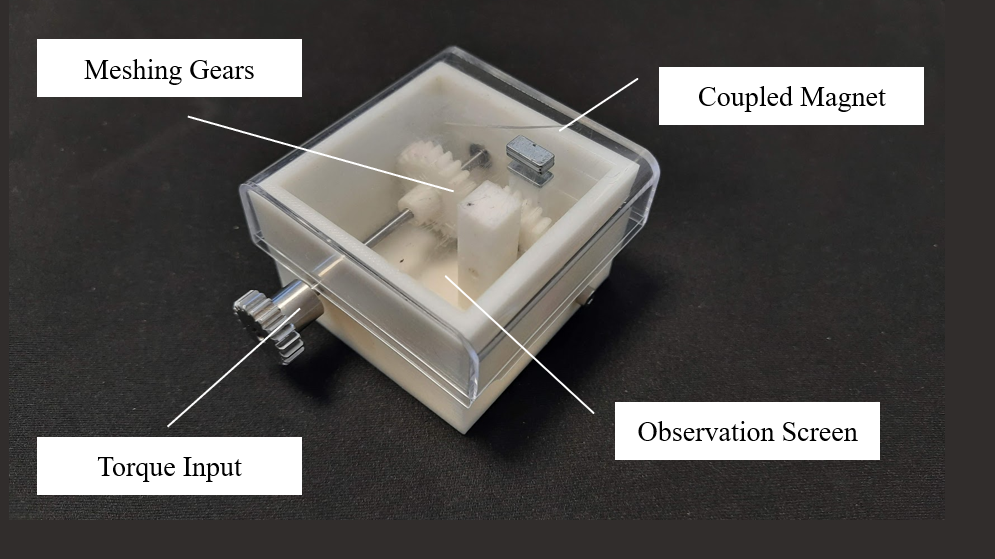}}
\caption{First prototype: scotch yoke mechanism.}
\label{fig1}
\end{figure}

The scotch yoke mechanism uses gears linked to a slot that oscillates back and forth as the gear rotates. We have fabricated a proof of concept as the first prototype seen in \figurename\ref{fig1}.

\begin{figure}[t]
\centerline{\includegraphics[width=3in]{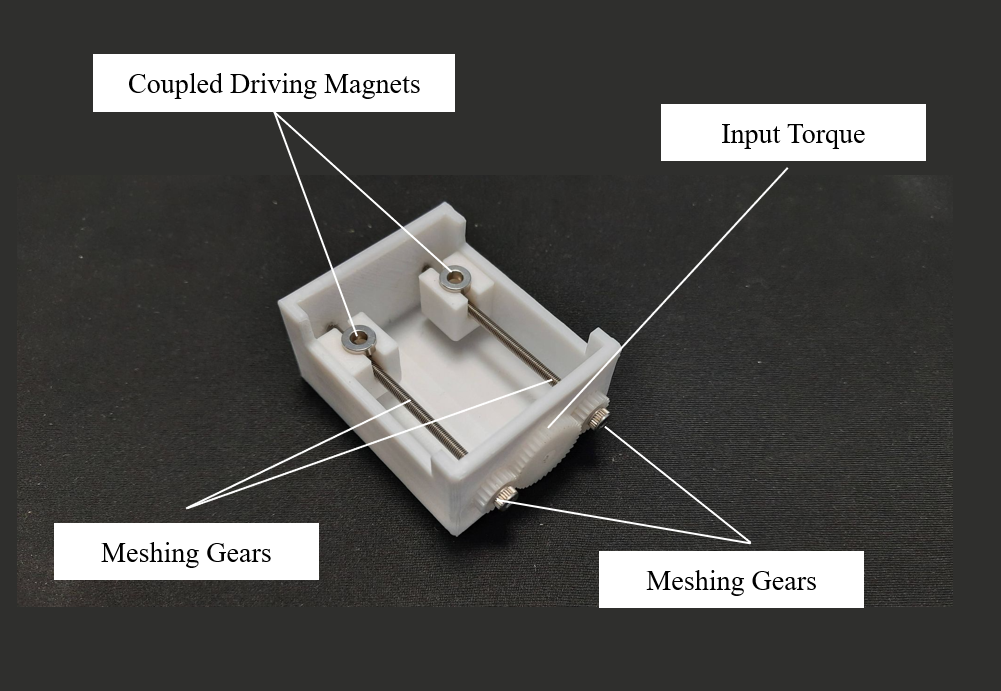}}
\caption{Second prototype: screw mechanism.}
\label{fig2}
\end{figure}

\begin{figure*}[!t]
\centering\includegraphics[width=7in]{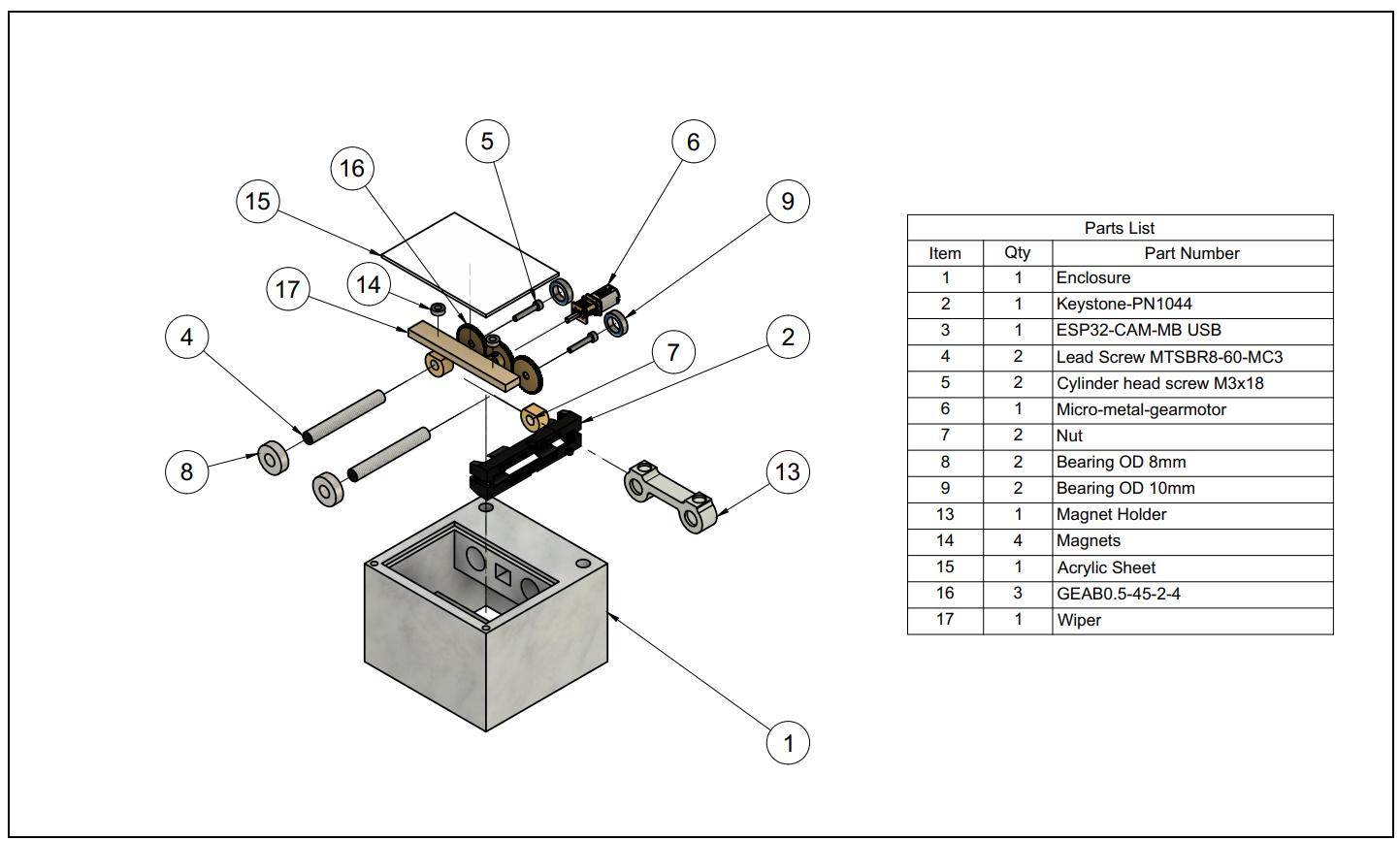}
\caption{Final prototype parts list.}
\label{fig3}
\end{figure*}

There was a great trade off of the scotch yoke mechanism encumbrance and its limited motion since the displacement of the wiper was constricted to the diameter of the gear. Hence, we have changed the mechanism to a more compact design seen in \figurename\ref{fig2} and \figurename\ref{fig3}. In this new mechanism which used screws, the torque input gear was coupled to two gears separately connected to two lead screws. This transformed the rotary to the linear motion. With this, the magnet holder enclosed the nuts that housed a pair of driving magnets. This moved the outer magnets parallel to its displacement. The length of the screw component determined the wiper coverage. The sensor was placed in the middle to avoid visual obstruction. When in operation, the magnet holder has resting positions on the far end sides.

The fabricated low-cost proof of concept mechanism as second prototype used polylactic acid filament using the Anycubic Mega Zero 2.0 3D Printer. A module 1.0 20-teeth resin gear was used to drive 38-teeth resin gears rigidly linked to M3x75 mm screws using tightened hexagonal M3 nuts are used to drive cubic magnet holders. 

\subsection{Built-in Electromechanical Wiper}

\begin{figure}[t]
\centerline{\includegraphics[width=3in]{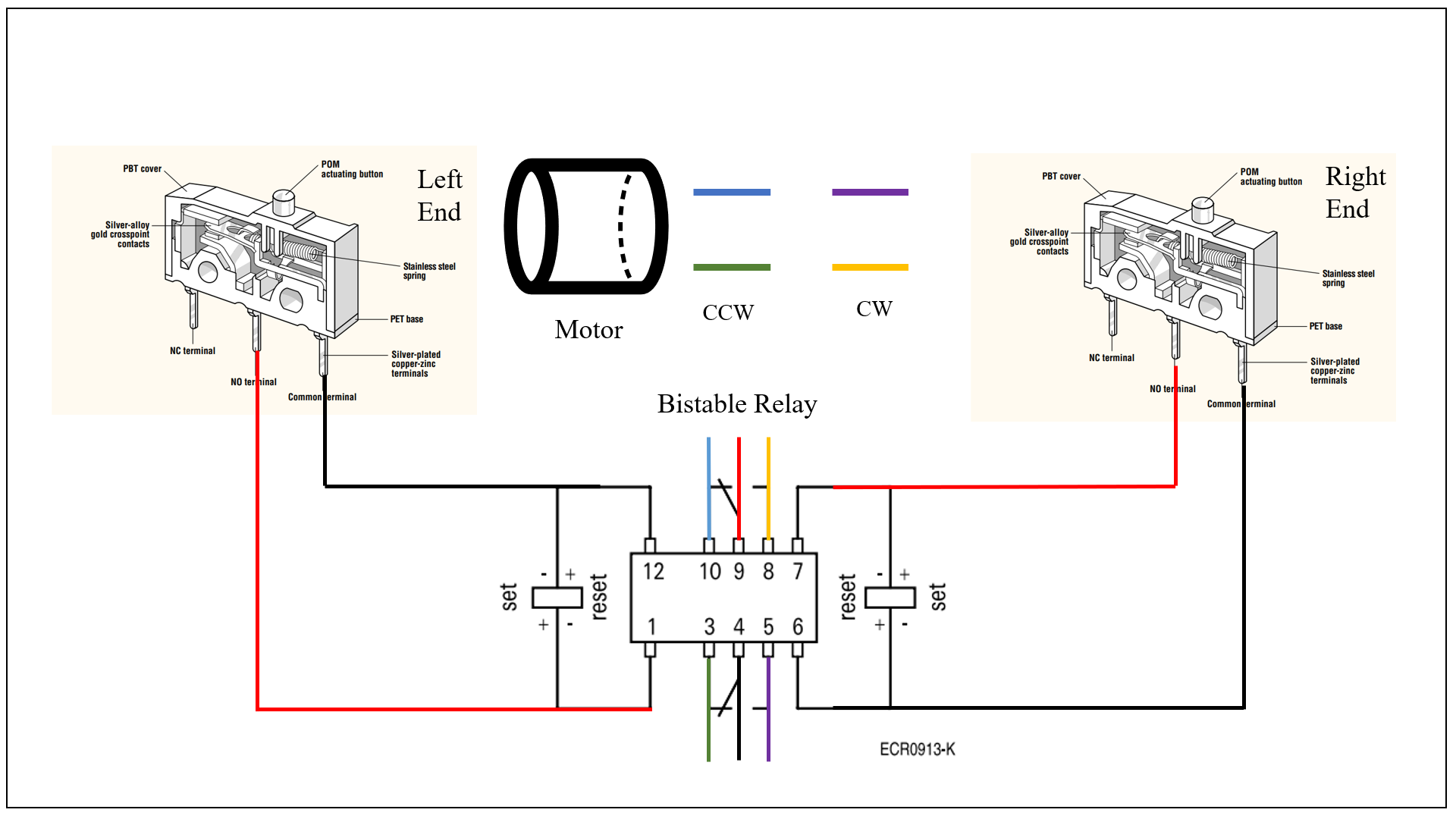}}
\caption{Schematic diagram of motor electronics.}
\label{fig4}
\end{figure}

A Tyco Electronics P2 relay (model V23079-B1001-B301) bistable, 2 coil latching relay was used to create a powered mechanism independent of a motor driver as a separate device. This mechanical electronic component enabled the forward and backward motion of the wiper, without the need of a driver in this proof of concept. This makes the mechanism easy to integrate with existing designs. The Omron D2F-L3-D3 limit switches were positioned in both ends of the wiper coverage, enabled the switching of the polarities of the source thus creating the cleaning motion. The electronic board schematic design can be seen in \figurename\ref{fig4}. The images of the components were taken from the official datasheet.

Cleaning times using the motor and the mechanism platform used for the experiment were recorded over 10 replicates. Average one cycle cleaning time was the mean time it took the wiper to traverse the whole coverage in one direction.
\subsection{Camera and Image Acquisition}

\begin{figure}[t]
\centerline{\includegraphics[width=3in]{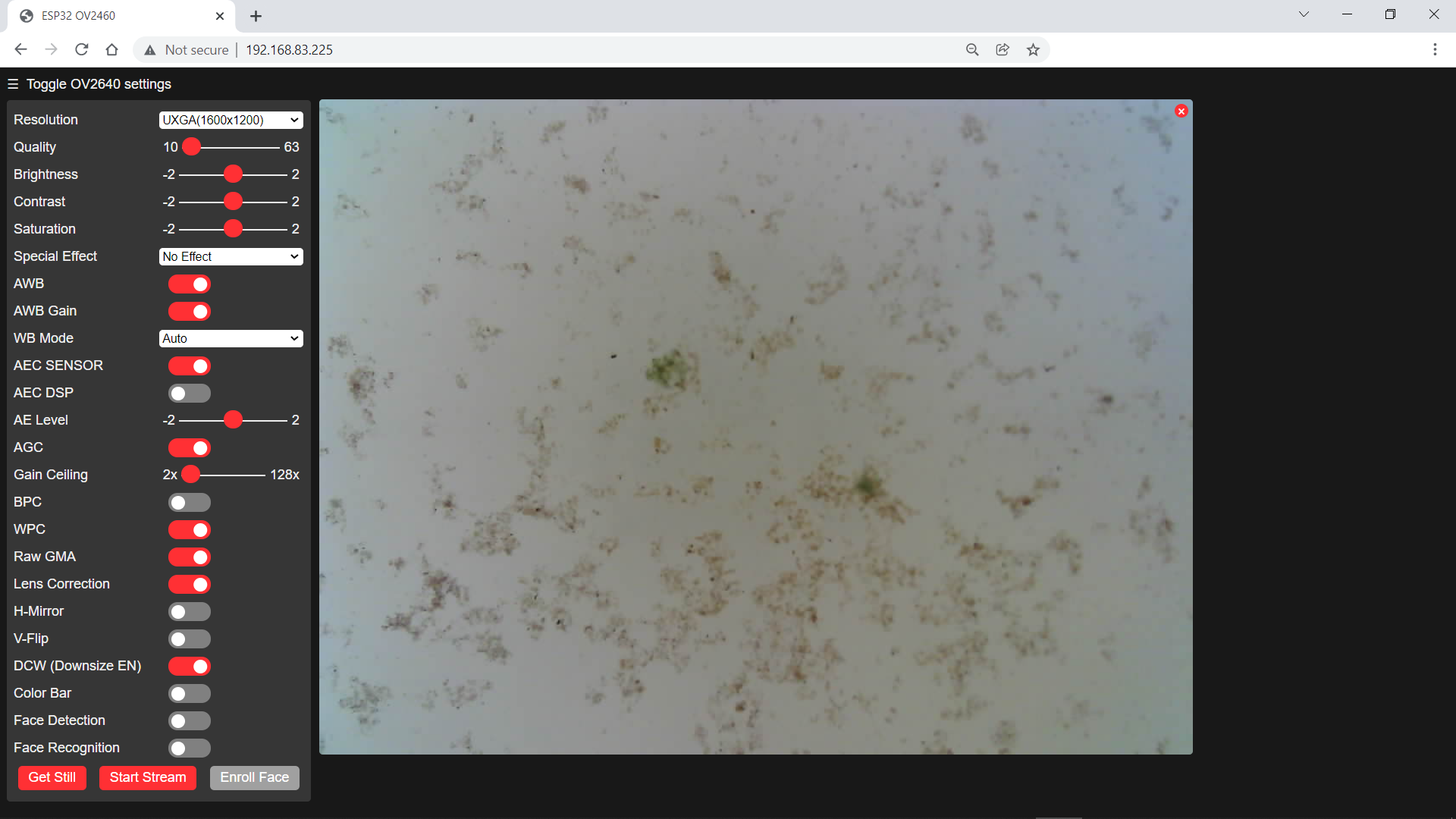}}
\caption{Webapp interface.}
\label{fig5}
\end{figure}

The low-cost ESP32-CAM is a small camera model with the ESP32-S chip features a microSD slot to store images and wifi module for connectivity. The built-in CameraWebServer webapp of ESP32-CAM, with interface shown in \figurename\ref{fig5} was used to stream the camera in real time while connected to a network and take the images. This allowed remote access to the image samples within the wifi coverage. It was programmed using an ESP32-CAM-MB Adaptor. The lens was manually adjusted to be focused on the bottom surface of the petri dish. 

\subsection{Experimental Design}
Experiments were conducted in a semi-field condition due to the unfavorable weather conditions for algae proliferation in the winter season. Water and algae samples were taken from a natural pond in Ponsacco, Pontedera, Pisa, Italy. Five milliliters of this sample water with the same amount of algae inoculations were each placed in six acrylic 120 mm-diameter petri dish containers. A liquid biostimulant based on alfalfa, algae and molasses were used to encourage proliferation. Ten milliliters of compound were diluted in 2 L of water. Three milliliters of this diluted solution were added to accelerate the growth. Components were further detailed in Table \ref{tab1}.

\begin{table}[htbp]
\caption{Biostimulant Components}
\begin{center}
\begin{tabular}{|c|c|}
\hline
\textbf{Component} &\textbf{Composition}\\
\hline
Organic Nitrogen& 2\%  \\
\hline
Organic Carbon of biological origin& 12\%  \\
\hline
Potassium Oxide soluble in water& 6\%  \\
\hline
Betaine& 1.18\%  \\
\hline
\end{tabular}
\label{tab1}
\end{center}
\end{table}

Conditions including temperature of 22-25°C and photo period of 16 hours of light and 8 hours of dark were kept constant in the observation room. 
\subsection{Data Analysis}
For each observation day (Day 0, 8, 13, 16), areas of observation screen with size 45x60 mm were sampled and accounted for the focal coverage of the sensor. A background made of white cardboard was placed 100 mm distance away from the plane of the petri dish and was used to lower the complexity of analysis. Three RGB replicate images were taken for the control sample before cleaning, and treated sample after cleaning. The 1600x1200 resolution images were taken with OV2640 Color CMOS UXGA 2.0 MegaPixel CameraChip™ with OmniPixel2™ Technology in ESP32-CAM and saved in an external computer. Images were then post-processed using Matlab R2020A™. The JPG RGB images were converted to grayscale and then thresholding was applied by using imbinarize, with sensitivity = 0.5, foreground polarity = dark and adaptive thresholding, to employ locally adaptive image threshold using first-order image statistics. Essentially, this step limited the values of the elements to either 0 and 1, given the calculated threshold. After which, the treated and control image qualities were then compared using a full reference metric called the mean squared error (MSE) (equation\ref{MSE}) between the image arrays \cite{b9}. The larger the value MSE, the better the effectiveness of the cleaning of the wiper. When compared to the reference as in Day 0, the lower the value of MSE, the more similar is the image quality.
\begin{equation}
\label{MSE}
MSE = \frac{1}{N} \sum\limits_{k=0}^N (X-Y)^2
\end{equation}
 where X, Y are vectors of image values and N is the number of elements.

\begin{figure}[t]
\centerline{\includegraphics[width=3in]{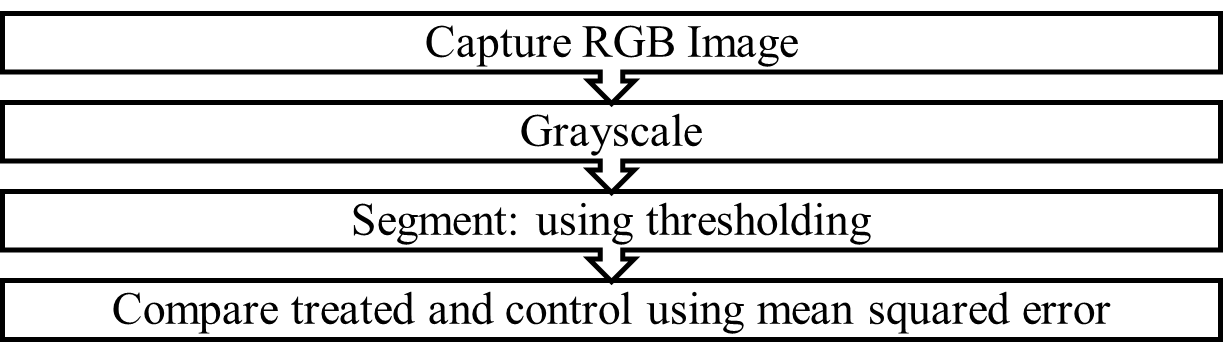}}
\caption{Data analysis flowchart.}
\label{fig6}
\end{figure}
\section{Results and Discussion}
\begin{figure}[t]
\centerline{\includegraphics[width=2in]{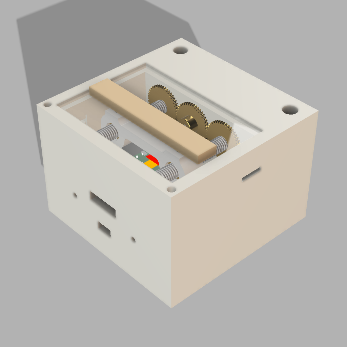}}
\centerline{\includegraphics[width=2in]{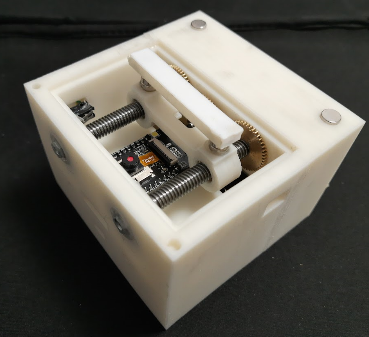}}
\caption{Final design rendering (top) and actual prototype (bottom).}
\label{fig7}
\end{figure}

In this paper we have made a magnetic-coupling based mechanism as antifouling system for underwater sensors. The design rendering can be seen with the actual prototype in the \figurename\ref{fig7}. 

The wiping unit was driven by the mechanism powered at rated voltage of 6V. This was done times full cycle back and forth to get the average stable power consumption of 0.84 Watts with 8 s average cleaning time in one direction summarized in Table \ref{tab2}. 

\begin{table}
\caption{Mechanism Characterisation}
\label{tab2}
\centering
\begin{tabular}{|c|c|}
\hline
One Pass Average Cleaning Time (seconds)	& 8 s \\
\hline
Rated Voltage (Volts) & 6.00 V \\
\hline
Steady State Current (Amperes)	& 0.14 A \\
\hline
Steady State Power (Watts)	& 0.84 W \\
\hline
Energy Consumption (Joules)	 & 6.72 J \\
\hline
\end{tabular}
\end{table}

\begin{figure}[t]
\centerline{\includegraphics[width=3in]{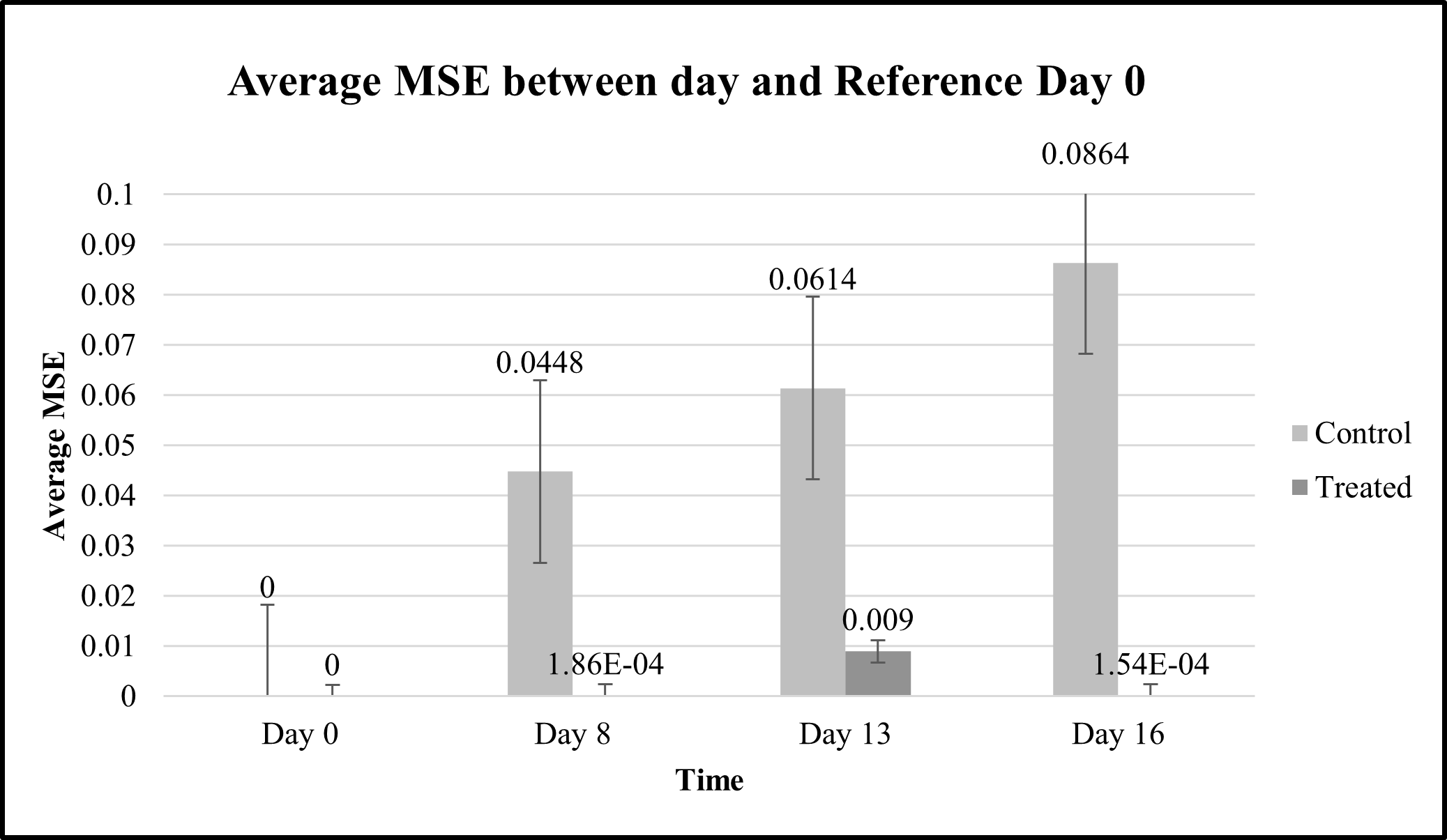}}
\caption{Average mean squared error between given day and reference Day 0.}
\label{fig8}
\end{figure}

The experiment was used to validate the effectiveness of the mechanism using the image sensor. The control and treated samples were compared using the MSE. When compared to the initial conditions, with the aid of stimulant to encourage algae growth, there was greater adhesion of fouling on the surface qualitatively. Both the control and treated samples were compared to the reference Day 0. The control samples were increasingly noisy until 0.0864 as seen in the graph in \figurename \ref{fig8}. Also, the treated samples have very similar in the image qualities with respect to the reference. This means that the image qualities were comparable after the wiping treatment. The error bars were the standard errors.

The increasing trend up to 0.0857 of average MSE between the control and treated samples per day, in comparison to the reference, signifies the effectiveness of the wiper technology. This means that even after 16 days of no antibiofouling intervention on the sensor unit, we can disrupt the biofouling and achieve a cleaned surface with comparable image quality to the initial state of observation. The error bars were the standard errors. This agrees with the previous work that the use of wiper helps prevent biofouling. Furthermore, the proposed mechanism may be a viable alternative to the current designs in the market today. 

\begin{figure}[t]
\centerline{\includegraphics[width=3in]{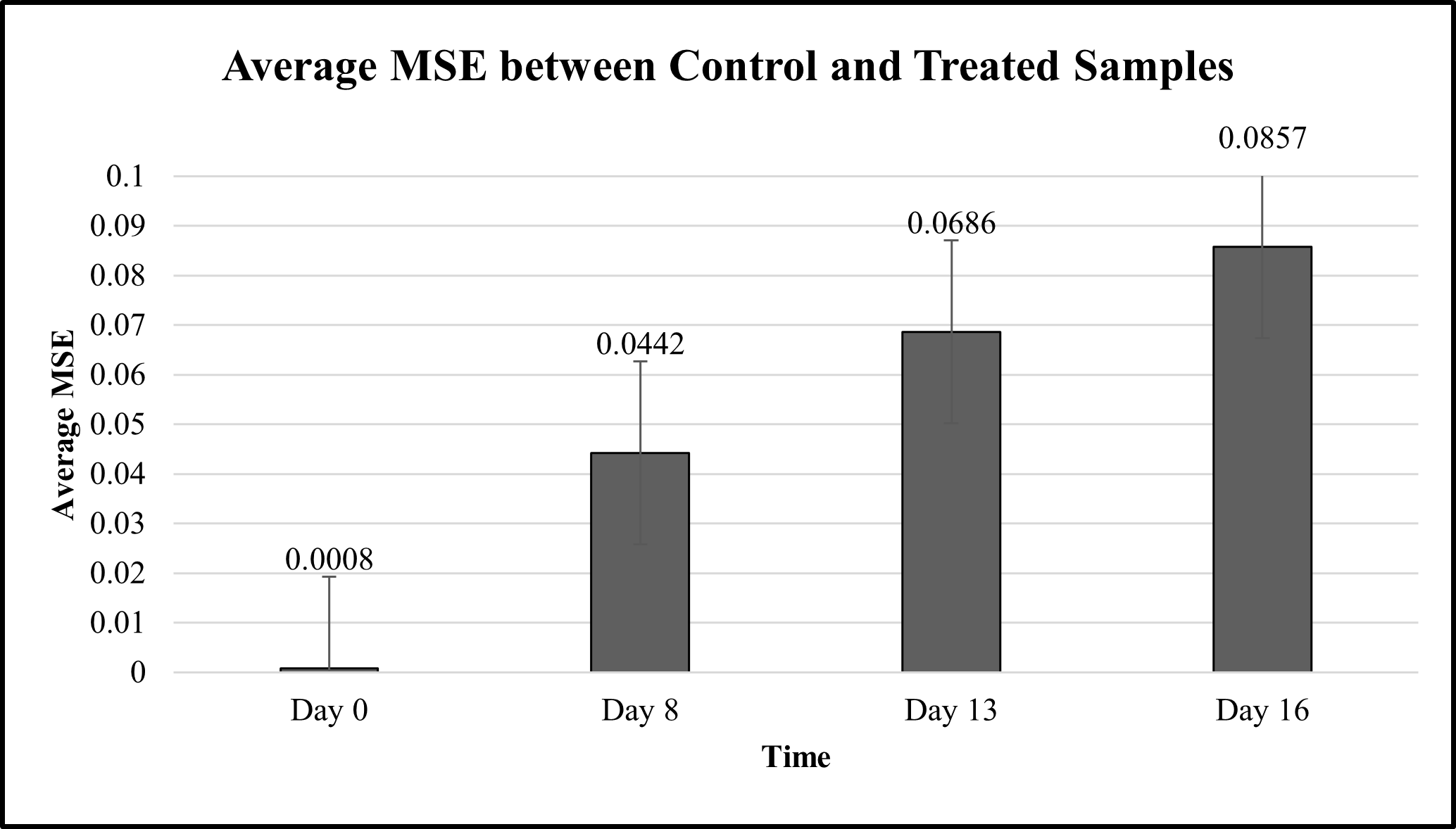}}
\caption{Average mean squared error between control and treated samples.}
\label{fig9}
\end{figure}
\begin{figure}[t]
\centerline{\includegraphics[width=2in]{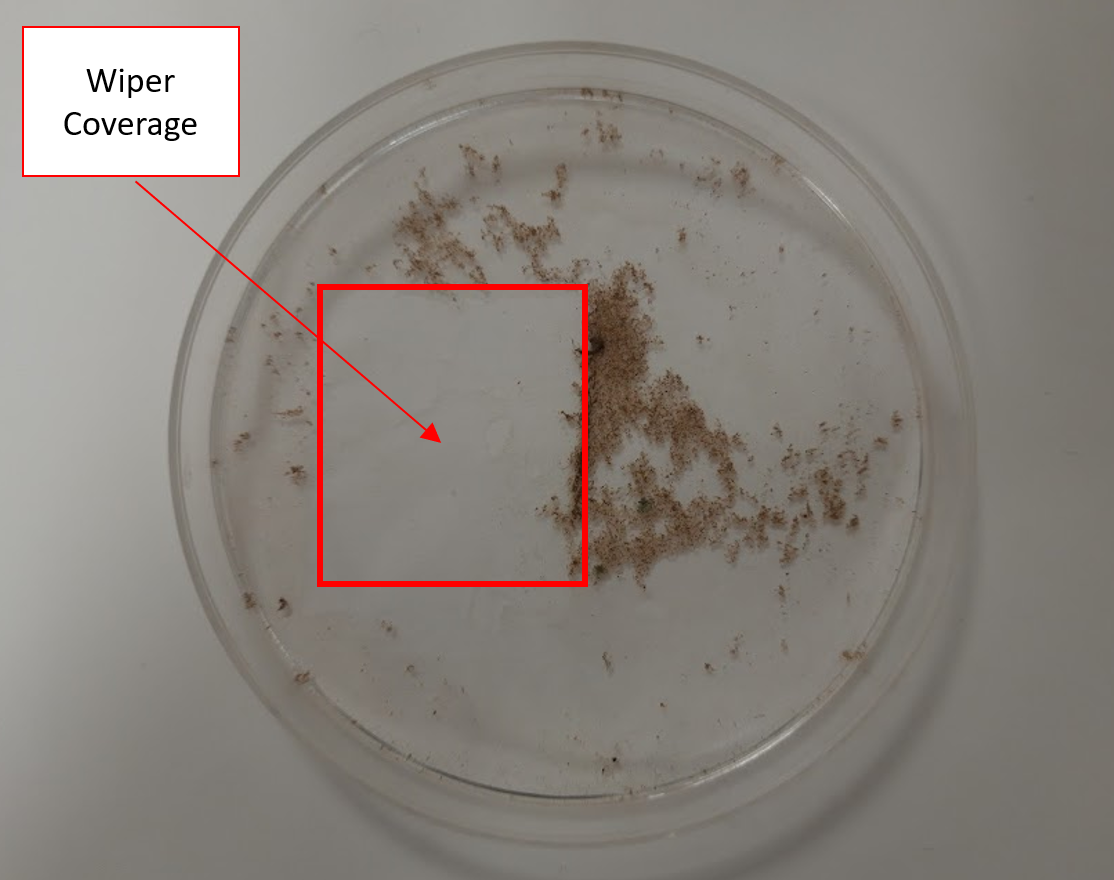}}
\caption{Image of Day 16 sample showing wiper coverage.}
\label{fig10}
\end{figure}

Semi-field conditions were appropriate because of weather limitations in the natural environment. Accounting for Day 16 of observation in \figurename\ref{fig10} shows the qualitative results of the wiper mechanism. As compared to the area outside the wiper coverage, we have greatly reduced the fouling using the proposed mechanism. 

Although, it has been observed that there was slight scratching on the surface of the observation screen after the wiping. This can be avoided by using a thicker material or by exploiting a different material. The cost may be reduced by using screws and plastic gears as in the second prototype. Of course, there is a trade-off in the quality and precision of the product.

Hence, we have proven the effectiveness of using magnetic coupling to remove biofouling quantitatively and qualitatively. When coupled with control, it can be used autonomously to remove noise from the optical sensor data. Since we have achieved great results after one pass of wiping within 16 days in the span of our observation, we did not proceed to test multiple wiping effectiveness. This experiment will need a longer time to grow the fouling.

\section{Conclusion}
In this paper we have designed, fabricated, and tested a configuration of a magnetic coupling-based antibiofouling mechanism using the components detailed in the previous sections. Based on 16 days of observation time with the wiping treatment, we have achieved to improve the image quality assessed with mean squared error by as much as 0.0857, out of 0.0864 baseline comparison with initial time of observation. Qualitatively, there’s a significant difference between the control and treated samples where the latter have a very similar quality as compared to reference samples in Day 0. We have found that it is effective after one pass of cleaning in 8 seconds with steady-state power of 0.84 W. The final prototype was low power, but not low cost if the components were not recycled. It is better to reuse existing parts to achieve the same motion, as in the first two prototypes. The proposed transmission of motion has been investigated and concluded effective in wiping biofilms and debris. Further, the use of autonomous activation based on image processing may likely be advantageous for minimizing the need for human inspection and maintenance in several applications. With a longer observation time, we can assess the effectiveness of multiple wiping configuration. Also, some materials used like in the non-abrasive cleaning unit can be improved to prevent scratching of the surface with prolonged use.
 
\section*{Acknowledgment}

Authors would like to thank all the members of Creative Engineering Laboratory for their support especially Godfried Jansen van Vuuren and Gaspare Santaera. J.P. Ramirez was funded by the Foreign Graduate Scholarship of the Philippine Department of Science and Technology Science Education Institute.

\vspace{12pt}

\end{document}